# Machine-Learning-Based Diagnostic Framework for Passive Ultrasonic Detection of Railway Wheel Defects

**Aashish Shaju[1], Steve Southward[1], and Mehdi Ahmadian[1]**

[1]Department of Mechanical Engineering, Virginia Tech
Blacksburg, VA 24061, USA
emails: aashish00@vt.edu, scsouth@vt.edu, ahmadian@vt.edu

## ABSTRACT

Reliable identification of wheel defects is crucial for railway safety and lowering maintenance costs. Passive air-coupled ultrasonic acoustic emission (UAE) sensing offers a promising non-contact alternative to traditional inspection methods, but its ability to distinguish between multiple defect types has not been fully explored. This work develops a machine-learning-based diagnostic framework that integrates nonparametric statistical testing, information-theoretic feature ranking, and supervised classification to assess the feasibility of multi-class defect identification using passive UAE data. Acoustic signals were collected from eleven MxV Rail wheelsets across nine health states, including healthy wheels, shattered rim cracks, rolling contact fatigue, spalls, blind holes, notches, and flange damage. A set of time- and frequency-domain features was evaluated with Kruskal-Wallis testing and mutual-information analysis to find the most discriminative features. A Random Forest classifier trained on the top features achieved a balanced accuracy of about 0.66 and a Macro-F1 score of 0.65 across all nine classes, with a small subset of just 3 to 4 features maintaining over 98% of the full-model accuracy. These findings establish a foundation for deployable, non-contact, machine-learning-enabled inspection tools and encourage further development toward field-ready passive UAE systems.

## INTRODUCTION

Railway wheel defects such as shattered rim cracks (SRC), rolling contact fatigue (RCF), spalls, flange damage, and other discontinuities pose significant risks to safe rail operations. In North America, RCF alone is estimated to cost more than 300 million USD annually in repairs and replacements [1]. These safety and economic concerns drive the ongoing development of diagnostic technologies capable of reliably detecting defects with minimal disruption to operations.

Conventional inspection techniques each face practical limitations for high-throughput deployment. Contact ultrasonic testing requires couplant and controlled surface conditions, making in-motion inspections impractical. Machine-vision systems are limited to surface-visible defects and are sensitive to environmental factors. Acoustic systems developed for wheel flats or bearing faults struggle to isolate crack-related ultrasonic emissions from background noise [2]. Guided-wave and laser-ultrasonic methods have demonstrated internal flaw detection capabilities, but field deployment is often hindered [3]. These challenges motivate the development of non-contact approaches capable of capturing diverse wheel defect signatures.

Recent work on passive air-coupled ultrasonic acoustic emission (UAE) sensing has established a validated acoustic fingerprint for at-rest wheel crack detection, demonstrating that the ultrasonic decay rate extracted from hammer-induced bursts provides strong separation between healthy and damaged wheels [4]. However, that study mainly focused on binary detection rather than distinguishing specific defect types. Parallel advances in machine learning for ultrasonic nondestructive evaluation have shown that tree-based ensembles and neural networks can extract meaningful diagnostic indicators from complex time-series data [5]. Building on this foundation, the present study develops an integrated framework combining advanced statistical testing, information-theoretic feature ranking, and supervised Random Forest classification to evaluate multi-class defect identification using passive UAE data.

## EXPERIMENTAL SETUP

Data was collected from a library of eleven full-scale railway wheelsets provided by MxV Rail (see Figure 1). The collection includes nine distinct health states: healthy wheels (H), shattered rim cracks (SRC), rolling contact fatigue (RCF), spalls (SP), blind holes (BH), notches (NT), flange damage (FL), and various combination conditions such as RCF+SP, RCF+SP+SRC, and SP+NT. During testing, the wheelsets were mounted on short rail sections to maintain realistic boundary conditions and acoustic coupling.

UAE bursts were generated using a standardized hammer impact applied at controlled locations on each wheel. Acoustic responses were recorded with two broadband air-coupled sensors (20-80 kHz) positioned in axial and radial orientations relative to the wheel web. Initial analyses showed consistently better class separation in the axial orientation due to clearer coupling paths and less masking from modal reflections; therefore, axial sensor data forms the basis for all subsequent analysis.

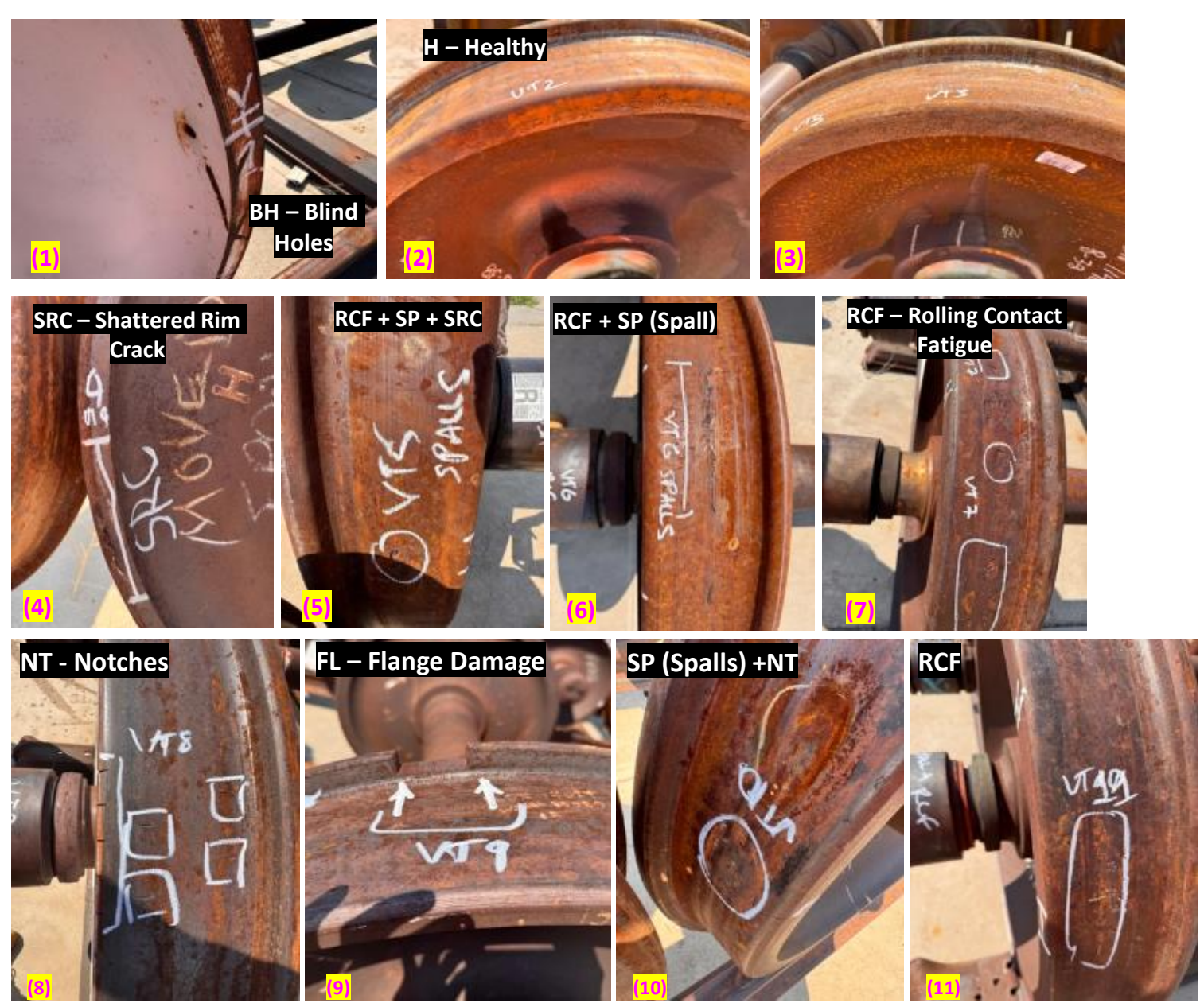

**Figure 1: Representative wheelsets from the test library showing (a) blind-hole defects, (b) healthy wheel, (c) shattered rim crack, and (d) rolling contact fatigue.**

Each recorded burst underwent automated preprocessing to identify high-energy transient segments, extract the signal envelope, and compute diagnostic features. Time-domain features included decay rate, RMS energy, and kurtosis. Frequency-domain features included spectral centroid, spectral standard deviation, and spectral entropy. These features serve as physically interpretable descriptors linked to damping behavior, frictional micro-slip, resonance shifts, and broadband energy distribution associated with different defect mechanisms.

# ANALYTICAL FRAMEWORK AND RESULTS

## Statistical Separability

The feasibility of multi-class defect discrimination was assessed using nonparametric statistical tests. For each feature, a Kruskal-Wallis H-test determined whether the distributions varied significantly across the nine defect classes. For features with significant overall differences, Dunn-Sidak post-hoc comparisons identified specific class pairs responsible for the separation. Several features exhibited strong class-level separability. The decay rate showed higher median values for friction-driven defects such as SRC and FL, aligning with the physical interpretation that crack interfaces increase high-frequency damping and that significant mass loss alters the fundamental vibrational mode. Conversely, non-frictional defects such as RCF had distributions overlapping with healthy wheels, indicating limited discriminatory ability from decay-based metrics alone and emphasizing the importance of a multi-feature classification approach.

## Information-Theoretic Feature Ranking

Mutual information (MI) was used to quantify each feature's diagnostic value by measuring how much it reduced uncertainty about the defect label based on the feature value. MI values were standardized using a permutation-derived z-score (z-MI) to allow fair comparison across features with different scales. The decay rate achieved the highest MI score (z-MI $> 20$), reaffirming its role as the leading discriminator. RMS energy, envelope low-frequency power, kurtosis, and skewness also scored highly, indicating that waveform damping, broadband energy distribution, and impulsivity all contribute significantly to defect-specific acoustic fingerprints. Stability analyses confirmed that MI rankings stayed consistent across different binning configurations, with Spearman rank correlations exceeding 0.9.

The combined MI ∩ Statistical heatmap in Figure 2 merges the two analysis stages into a single, decision-focused visualization. Each square represents a pair of defect classes, with its color indicating the feature that (1) shows statistically significant separation between classes (Kruskal–Wallis and Dunn–Sidak) and (2) has the highest mutual information among the qualifying features. The colored circles inside each square display all features meeting the statistical significance threshold for that pair, allowing the viewer to see when multiple features are viable candidates, while the background color emphasizes the most informative one. This representation highlights clear patterns. Features such as Decay Rate, Kurtosis, Skewness, and Envelope Low-Frequency Power consistently appear across many defect pairs, affirming their robustness and importance as key discriminative features for the downstream classifier. In contrast, gray cells indicate defect pairs with no statistically significant separation or low MI, suggesting cases where passive UAE alone may be insufficient and additional sensing or refined features are needed. Overall, this visualization combines statistical and information-theoretic insights to aid in selecting a compact yet effective feature set for multi-class classification.

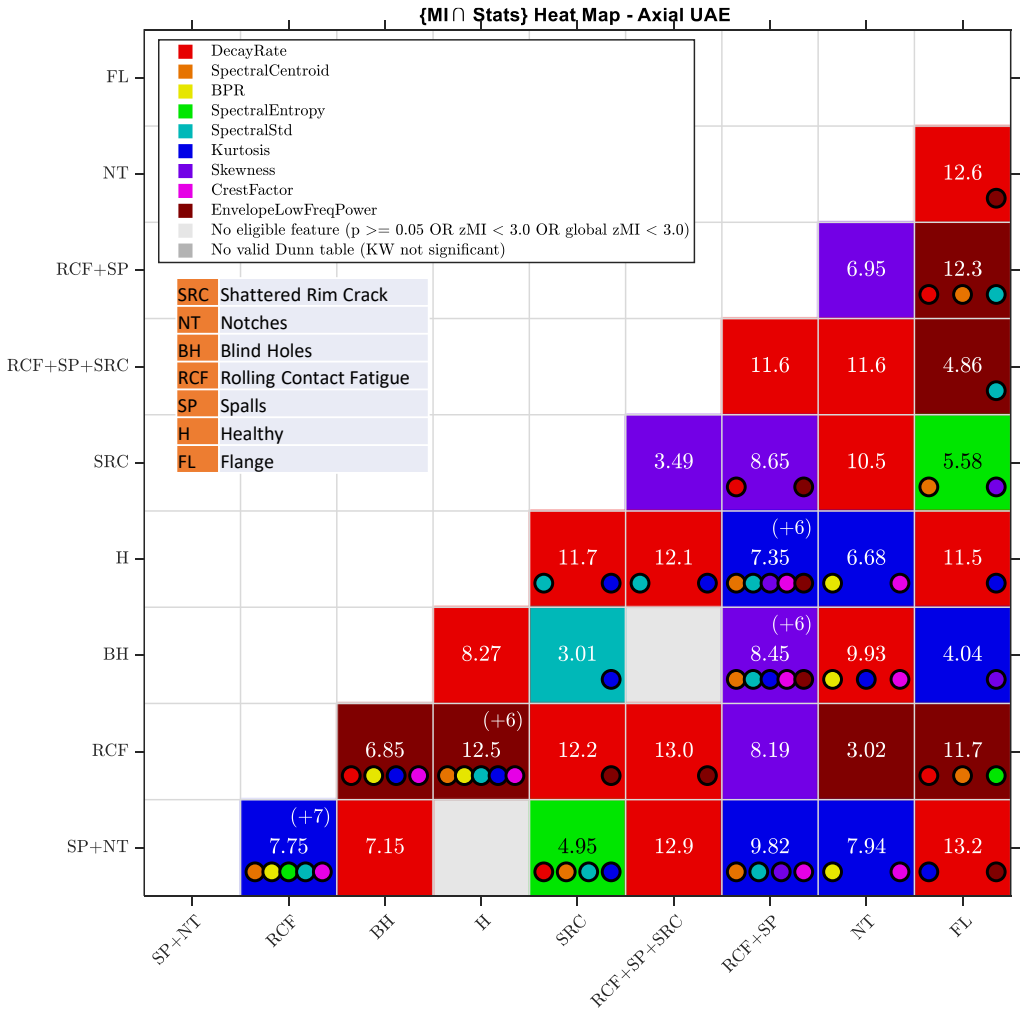

**Figure 2: Combined MI and statistical significance heatmap identifying the most informative and statistically separable feature for each defect-pair comparison.**

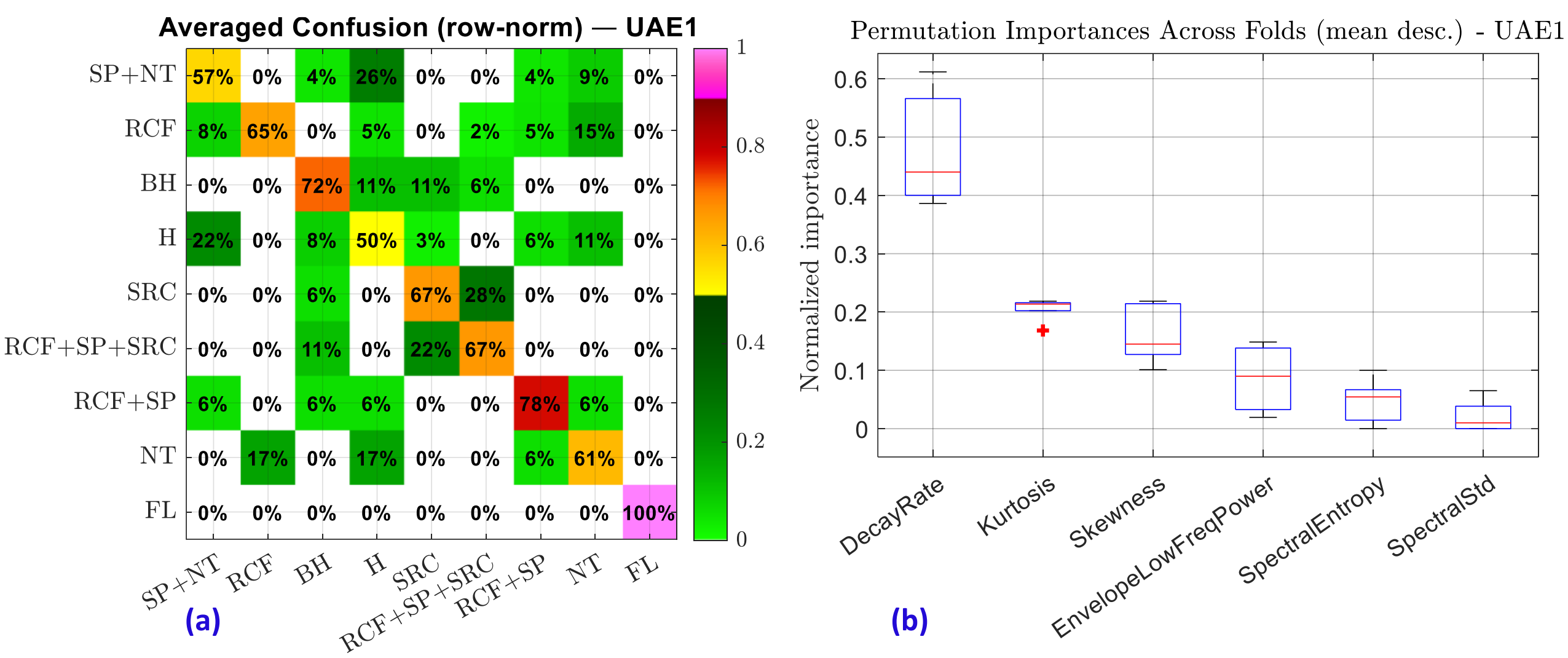

**Figure 3: (a) Row-normalized confusion matrix showing multi-class classification performance. (b) Permutation importance distribution across cross-validation folds.**

## Random Forest Classification

A Random Forest classifier with 500 trees was trained using stratified 5-fold cross-validation on the selected feature set. Fold-specific z-score normalization was applied to prevent data leakage, and inverse-frequency class weights were used to address class imbalance.

The row-normalized confusion matrix in Figure 3(a) summarizes the multi-class classification performance across the health states of nine wheels. The model achieved a balanced accuracy of approximately 0.66 and a Macro-F1 score of 0.65. High recognition rates were obtained for FL (100%), RCF+SP (78%), BH (72%), and SRC (67%). Moderate performance was observed for RCF (65%), NT (61%), and RCF+SP+SRC (67%), while lower accuracy was seen for H (50%) and SP+NT (57%). Most misclassifications occur between physically similar conditions, such as healthy wheels being confused with NT or SP+NT, which aligns with their relatively weak acoustic signatures.

Feature importance analysis based on permutation importance across cross-validation folds is shown in Figure 3(b). The results indicate that the **decay rate** is the most influential feature, followed by **kurtosis**, **skewness**, and **envelope low-frequency power**. The consistency of these rankings across folds suggests that a small set of physically meaningful features captures most of the discriminative information needed for classification.

## CONCLUSION AND FUTURE WORK

This study developed and validated a machine-learning-based diagnostic framework for multi-class identification of railway wheel defects using passive air-coupled UAE data. The key findings are threefold. First, nonparametric statistical analysis confirmed that different defect types produce measurably distinct and separable acoustic signatures, establishing a solid physical basis for automated classification. Second, information-theoretic evaluation showed that features such as decay rate, kurtosis, skewness, and envelope low-frequency power consistently provide the most mutual information and remain stable across different configurations. Third, the Random Forest classifier trained on this optimized feature set achieved a balanced accuracy of 0.66 and a Macro-F1 score of 0.65, with physically interpretable confusion patterns confirming structured, non-random performance. Permutation importance analysis further confirmed that a small number of time-domain features, led by decay rate, dominate the classifier's ability to discriminate, supporting the potential for computationally efficient, real-time classification suitable for field deployment.

Future work will focus on three main areas. First, enhancing model accuracy through improved feature engineering, such as wavelet and higher-order spectral features, along with more powerful tree-based learners like Gradient Boosted Trees (XGBoost, LightGBM). Second, developing a near real-time version of the optimized model on a portable handheld device to enable in-field UAE data collection and expand the training dataset. Third, evolving the mature system into a compact wayside prototype that uses naturally occurring wheel-rail excitation at rail discontinuities to evaluate the feasibility of continuous, non-contact wheel-health monitoring in operational environments.

## ACKNOWLEDGMENTS

This research was funded by MxV Rail, grant number 453624. The authors sincerely thank Drs. Anish Poudel and Corey Pasta for many valuable technical discussions that contributed to the direction and strength of the analysis in this study.